\documentclass[10pt,twocolumn,letterpaper]{article}

\usepackage[pagenumbers]{cvpr} % To force page numbers, e.g. for an arXiv version

\usepackage{graphicx}
\usepackage{amsmath}
\usepackage{amssymb}
\usepackage{booktabs}
\usepackage{times}
\usepackage{epsfig}
\usepackage{tabu}
\usepackage{color}
\usepackage{multirow}
\usepackage{algorithmic}
\usepackage{bbding}
\usepackage{colortbl}
\usepackage{multicol}
\usepackage[ruled, lined, linesnumbered, commentsnumbered, longend]{algorithm2e}

\makeatletter
\def\hlinew#1{%
  \noalign{\ifnum0=`}\fi\hrule \@height #1 \futurelet
   \reserved@a\@xhline}
\makeatother

\definecolor{mygray}{gray}{.9}

\usepackage[pagebackref,breaklinks,colorlinks]{hyperref}

\usepackage[capitalize]{cleveref}
\crefname{section}{Sec.}{Secs.}
\Crefname{section}{Section}{Sections}
\Crefname{table}{Table}{Tables}
\crefname{table}{Tab.}{Tabs.}

\def\cvprPaperID{6758} % *** Enter the CVPR Paper ID here
\def\confName{CVPR}
\def\confYear{2022}

\begin{document}

%%%%%%%%% TITLE - PLEASE UPDATE
\title{Push Stricter to Decide Better: A Class-Conditional Feature Adaptive Framework for Improving Adversarial Robustness}

\author{
    %Authors
    % All authors must be in the same font size and format.
    Jia-Li Yin\textsuperscript{\rm 1, 2},
    Lehui Xie\textsuperscript{\rm 1, 2},
   Wanqing Zhu\textsuperscript{\rm 1, 2},
     %\thanks{Corresponding author.},
    Ximeng Liu\textsuperscript{\rm 1, 2},
    Bo-Hao Chen\textsuperscript{\rm 3}
    \\
    \textsuperscript{\rm 1}Fuzhou University ~~~~\textsuperscript{\rm 2}NSIS Lab ~~~~\textsuperscript{\rm 3}YuanZe University\\
     {\tt\small jlyin@fzu.edu.cn, \{xsplend, zhuuwuuu, snbnix\}@gmail.com, bhchen@saturn.yzu.edu.tw}
}

\maketitle

%%%%%%%%% ABSTRACT
\begin{abstract}
In response to the threat of adversarial examples, adversarial training provides an attractive option for enhancing the model robustness by training models on online-augmented adversarial examples. However, most of the existing adversarial training methods focus on improving the robust accuracy by strengthening the adversarial examples but neglecting the increasing shift between natural data and adversarial examples, leading to a dramatic decrease in natural accuracy. To maintain the trade-off between natural and robust accuracy, we alleviate the shift from the perspective of feature adaption and propose a Feature Adaptive Adversarial Training (FAAT) optimizing the class-conditional feature adaption across natural data and adversarial examples. Specifically, we propose to incorporate a class-conditional discriminator to encourage the features become (1) class-discriminative and (2) invariant to the change of adversarial attacks. The novel FAAT framework enables the trade-off between natural and robust accuracy by generating features with similar distribution across natural and adversarial data, and achieve higher overall robustness benefited from the class-discriminative feature characteristics. Experiments on various datasets demonstrate that FAAT produces more discriminative features and performs favorably against state-of-the-art methods. Codes are available at \url{https://github.com/VisionFlow/FAAT}.
\end{abstract}

%%%%%%%%% BODY TEXT
\section{Introduction}
Deep feed-forward networks have brought great advances to the state-of-the-art across various machine-learning tasks and applications. However, these advances in performance can be dramatically degraded when faced with maliciously crafted perturbations, i.e., \textit{adversarial examples}. This phenomenon has raised growing concerns over the reliability of deep feed-forward networks and significantly hinders the deployment of these state-of-the-art architectures in practical~\cite{64xie2017adversarial,Sayles_2021_CVPR,Hendrycks_2021_CVPR,Wu_2021_CVPR,Pony_2021_CVPR}. 

\begin{figure}[!t]
\centering
\includegraphics[width=1.0\linewidth]{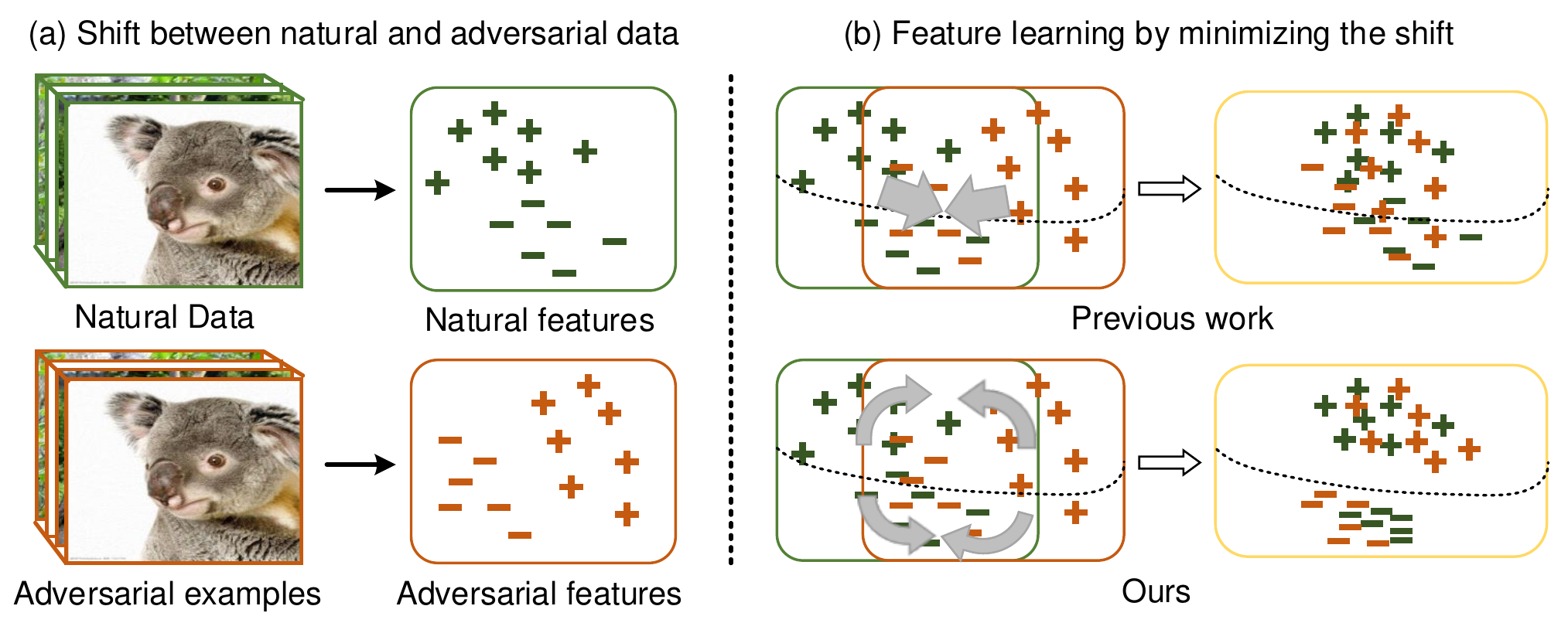}
    \caption{%Comparison between previous methods and ours. \textbf{Left}: Shift between natural data and adversarial examples. \textbf{Middle}: Previous methods focus on minimizing the feature discrepancy between natural data and adversarial examples. \textbf{Right}: Our method perform based on soft labels.
    Illustration for our method. (a) Shift between natural and adversarial data. Note that the adversarial examples are generated by adding imperceptible perturbations over the natural data. Human are not sensitive to the perturbations but CNN models can be easily affected as there is a clear shift between natural and adversarial feature distributions. (b) The comparison of previous methods and ours. Previous methods focus on minimizing the feature discrepancy between natural data and adversarial examples. Our method take class-level information into consideration for fine-grained feature adaption.}
    \vspace{-15pt}
\label{fig:introduction}
\end{figure}

In response to the threat of adversarial attacks, learning to increase model robustness by augmenting training data with adversarial examples is known as adversarial training. A number of approaches to adversarial training has been suggested in a formulation of min-max optimization, e.g., generating on-line adversarial examples that can maximize the model loss, and then the models are trained on these adversarial examples to minimize the loss, so that the model can prefer robust features when given adversarial data. The appeal of adversarial training approaches is to encourage the model to learn the distributions of adversarial data. This strategy can effectively improve the model robustness and has become the most popular way to defend adversarial examples. At the same time, however, these improvements of robustness against adversarial examples typically come at the cost of decreased natural data accuracy as there is a shift between adversarial and natural data distribution. Especially when previous works~\cite{madry2018towards, zheng2020efficient} focus on maximizing the attack strength of on-line generated adversarial examples to improve robustness, these methods work poorly on natural data, as reported in Table\,\ref{tab:robust}.

\begin{table*}[!t]
    
    \centering
    %\fontsize{9}{11.5}\selectfont
    \caption{Comparison of previous methods and our method under PGD-20/CW-20 attack for WideResNet-34-10 on MNIST, CIFAR-10, and CIFAR-100, respectively. $\text{Acc}_n$ presents accuracy on natural images while $\text{Acc}_{r1}$ and $\text{Acc}_{r2}$ denotes the robust accuracy against PGD-20 and CW-20 attacks. We highlight the higher accuracy in \textcolor{blue}{blue} and the lower in \textcolor{red}{red} compared to the standard adversarial training.}
    \label{tab:robust}
        \begin{tabular}{l|c|ccc|ccc|ccc}
        %\tabucline[1pt]{--}
        \hlinew{1pt}
        \multirow{2}{*}{Denfese Method} &
         \multirow{2}{*}{Trade-off term} &
         \multicolumn{3}{c|}{{MNIST}} &
          \multicolumn{3}{c|}{{CIFAR-10}} &
          \multicolumn{3}{c}{{CIFAR-100}} \\ 
         &&$\text{Acc}_n$&
          $\text{Acc}_{r1}$ &
          $\text{Acc}_{r2}$ &
         $\text{Acc}_n$ &
         $\text{Acc}_{r1}$ &
          $\text{Acc}_{r2}$ &
          $\text{Acc}_n$ &
         $\text{Acc}_{r2}$ &
          $\text{Acc}_{r2}$ \\ \hlinew{1pt}
          
               Baseline  & \XSolidBrush & 99.50       & 0.07 & 0.00 & 95.80 & 0.00 & 0.00  & 78.76& 0.00  & 0.00 \\ \hline
                Standard (PGD-10)~\cite{madry2018towards} & \XSolidBrush & 99.41   & 96.01 & 95.87 & 85.23 & 48.93 & 48.74  & 60.29& 26.84  & 26.25 \\
                 +ATTA~\cite{zheng2020efficient} & \XSolidBrush    & \textcolor{blue}{99.60}    & \textcolor{blue}{98.20} & \textcolor{blue}{97.72} & \textcolor{red}{83.30} & \textcolor{blue}{54.33} & \textcolor{blue}{54.25}  & \textcolor{red}{55.09}& \textcolor{red}{23.23}  & \textcolor{red}{22.85} \\
                +Trades ($\lambda=6$)~\cite{Zhang2019tradeoff}  & \Checkmark & \textcolor{blue}{99.48}     & \textcolor{blue}{96.50} & \textcolor{red}{95.79} & \textcolor{red}{82.64} & \textcolor{blue}{51.14} & \textcolor{blue}{50.93}  & \textcolor{blue}{62.37}& \textcolor{red}{25.31}  & \textcolor{red}{24.53}\\
                +ALP~\cite{58kannan2018adversarial}  & \Checkmark & \textcolor{red}{99.27} &\textcolor{blue}{97.80} &\textcolor{blue}{97.26} &\textcolor{red}{84.01} &\textcolor{blue}{54.96} & \textcolor{blue}{54.71} &\textcolor{red}{55.60}& \textcolor{blue}{28.28}  & \textcolor{red}{24.40} \\ 
                +Ours  & \Checkmark &\textcolor{blue}{\textbf{99.70}} & \textcolor{blue}{\textbf{98.53}} & \textcolor{blue}{\textbf{97.89}} &\textcolor{blue}{\textbf{86.74}} & \textcolor{blue}{\textbf{55.18}} & \textcolor{blue}{\textbf{54.79}} & \textcolor{blue}{\textbf{62.58}}&\textcolor{blue}{\textbf{30.22}}  & \textcolor{blue}{\textbf{30.58}}\\ \hlinew{1pt} 
        
        \end{tabular}
        
    \end{table*}

In this paper, we are working on alleviating the shift between natural and adversarial data distribution. There have been some pioneering works trying to address this problem. Kannan et al.~\cite{58kannan2018adversarial} proposed adversarial logit pairing (ALP) method to encourage the logits from original inputs and adversarial examples in the CNN model to be similar. Zhang et al.~\cite{Zhang2019tradeoff} provided a regularizer, namely Trades, to minimize the KL-divergence between the features of original inputs and adversarial examples. Cui et al.~\cite{cui2020learnable} proposed to parallelly train two models using original input and adversarial examples separately while the natural model is used to guide the training of robust model. Despite the improvements, these methods have two major issues. First, these methods generally employ arbitrary point-to-point metrics to evaluate the distribution discrepancy which do not guarantee the well distribution alignment. Second, there exist many feature points near the decision boundary, they may overfit the natural data but are less discrminative for the adversarial examples, as shown in Figure\,\ref{fig:introduction}.

%To address the above issues, we introduce a new Feature Adaptive Adversarial Training (FAAT) to enable trade-off between natural and robust accuracy for adversarial training. 
Unlike the previous methods, we propose to alleviate the shift from the perspective of feature adaption. To address the above issues, we focus on learning features that are both class-level discriminative and invariant to the change of adversarial attacks, i.e., learning the same or very similar distributions across the natural and adversarial data with same class. In this way, the obtained network can maintain the accuracy on both natural and adversarial domains. It is conceptually similar to the problem of domain adaption~\cite{Haoran_2020_ECCV, Ganin2015dann,kang2019contrastive}, which aims to alleviate the domain shift problem by aligning the feature distributions of different domains. Recent success in domain adaption lies in the incorporation of a discriminator to model the feature distribution, where the discriminator aims to distinguish the features from different domains while the trained network tries to fool the discriminator, so the features become domain-invariant.
%In domain adaption, a discriminator with hierarchical architecture can effectively model the feature distribution. Recent studies in domain adaption have applied the discriminator into domain adaption frameworks. The discriminator is these frameworks works to distinguish the features from different domains, then the trained network tries to fool the discriminator by generating domain-invariant features. first propose to incorporate such a discriminator into adversarial training framework to 

Motivated by this, we take advantage of such a discriminator in coming up with a novel adversarial training framework: Feature Adaptive Adversarial Training (FAAT), where a discriminator is incorporated to encourage the similar distribution of across natural data and adversarial examples. However, the simple discriminator only encourages distribution similarity across domains but neglects the underlying class structures. To make the features near the decision boundary more discriminative, the feature discrepancy among classes should be further ensured. Inspired by~\cite{Haoran_2020_ECCV}, we propose to directly encode class knowledge into the discriminator and encourage class-level fine-grained feature adaption of adversarial features. Specifically, the output of discriminator is transferred from binary domain label into class-conditional domain label. Such class-conditional discriminator can well assist the fine-grained feature adaption across natural data and adversarial examples according to their corresponding classes. 
%We conducted extensive experiments on diverse datasets. The experimental results show that the proposed method achieves impressive improvements. For example, as can be seen in Table\,\ref{tab:robust}, our proposed method improves the natural and PGD-20 robust accuracy of WideResNet trained with standard adversarial training from $60.29\%$ to $63.25\%$ and $26.84\%$ to $30.25\%$ on CIFAR-10 dataset. 
In summary, the proposed FAAT has the following advantages:
\begin{itemize}
    \item FAAT can effectively maintain the trade-off between natural and robust accuracy by encouraging feature adaption across natural data and adversarial examples.
    \item FAAT can significantly improve the overall robustness of adversarial training by incorporating class knowledge into feature adaption.
    \item FAAT provides an universal framework for adversarial training which can be easily combined with other methods.
\end{itemize}

%therefore propose to incorporate a domain discriminator into the adversarial training that can discriminates between adversarial and natural domain. 

%It is thus natural to wonder whether we can develop a training method that can achieve both high robust and natural accuracy. To answer this question, we first take a close look at the feature generation for adversarial and natural data during adversarial training process. 

%------------------------------------------------------------------------
\section{Related Work}
\subsection{Adversarial attacks}
An adversarial example refers to a maliciously crafted input that is imperceptible to humans but can fool the machine-learning models. It is typically generated by adding a perturbation $\delta$ subject on constraint $\mathcal{S}$ to the original input $x$, i.e., $x^{adv} = x+\delta$. Szegedy et al.~\cite{Szegedy} first observed the existence of adversarial examples and used a box-constrained L-BGFS method to generate $\delta$. Goodfellow et al.~\cite{Goodfellow2015} investigated the linear nature of networks and proposed the fast gradient sign method (FGSM) to generate more effective adversarial examples. Following this work, various attacks are proposed to generate stronger $\delta$ that can maximize the model training loss. Specifically, Madry et al.~\cite{madry2018towards} divided one-step FGSM into iterative multiple small steps to enhance the attack strength, known as Projected Gradient Descent (PGD) attack. Carlini et al.~\cite{cwattack} proposed CW attack based on the margin loss to enhance the attack. Croce et al.~\cite{croce2020reliable} ensemble multiple attacks into a more powerful attack, namely Auto-Attack (AA), which is known as the strongest attack so far.

An intriguing property of adversarial examples is that they can be transferred to attack different network architectures~\cite{Goodfellow2015,Szegedy2014intriguing}. Liu et al.~\cite{liu2017delving} first analyzed the transferability of adversarial examples and used ensemble-based methods to generate adversarial examples for attacking unknown models. Dong et al.~\cite{dong2018boosting} took the momentum information of the gradient into process and achieved more transferable adversarial examples. Xie et al.~\cite{Xie2019transfer}, Wu et al.~\cite{Wu_2020_CVPR} and Wang et al.~\cite{Wang_2021_transfer} further boosted the transferability of adversarial examples via creating diverse input patterns, attention scheme and invairance tuning, respectively, making the adversarial defense more challenging.

\subsection{Adversarial defenses}
Various defense methods against adversarial attacks have been proposed over the recent years, including detection-based methods~\cite{guo2017countering,buckman2018thermometer,63huang2019model}, input transformation-based methods~\cite{Liao2018denoise}, and training-based methods~\cite{Goodfellow2015,shafahi2019adversarial,wong2020fast,madry2018towards,tramer2017ensemble,zheng2020efficient,Zhang2020ES,Cai2018Curriculum}. Here we focus on the most related and effective one: adversarial training methods. Adversarial training performs on-line defending which aims to minimize the loss of model on on-line generated adversarial examples that maximize the model loss at each training epoch. This process can be described as:
	\begin{align}\label{eq1}
		\min _{\theta} \mathbb{E}_{(x, y) \sim \mathcal{X}}[\max _{\delta} \mathcal{L}(x+\delta, y; {\theta})],
	\end{align}
where $(x,y)$ is a training pair in dataset $\mathcal{X}$, $\mathcal{S}$ denotes the region within the $\epsilon$ perturbation range under the $\ell_{\infty}$ threat model for each example, i.e., $\mathcal{S}=\{\delta:\|\delta\|_{\infty} \leq \epsilon\}$, where the adversary can change input coordinate $x_{i}$ by at most $\epsilon$. It formulates a game between adversarial examples and model training. The stronger adversarial examples are generated; more robust model can be achieved. Thus existing adversarial training focus on maximizing the attack strength of adversarial training. Madry et al.~\cite{madry2018towards} used PGD to approximate the adversarial attack for adversarial training. Cai et al.~\cite{Cai2018Curriculum} and Zhang et al.~\cite{Zhang2020ES} further incorporated curriculum attack generation and early stopping into the PGD adversarial training to make it more efficient and practical. On the other hand, Tramer et al.~\cite{tramer2017ensemble} proposed ensemble adversarial training where the adversarial examples are generated by multiple ensemble models to enhance the attack strength. Several methods are proposed to further promote the diversity in the ensemble models, including output logits diversification~\cite{pang2019improving}, gradient direction minimization~\cite{kariyappa2019improving}, and vulnerability diversity maximization~\cite{yang2020dverge}. These methods can effectively improve the robustness of models but also leads to the decrease of natural accuracy due to the ignorance of the shift between natural and adversarial data distribution.

\subsection{Trade-off between natural and robust accuracy}
Besides simply pursuing the improvement over robustness, a recent line of work investigates the trade-off between natural and robust accuracy~\cite{58kannan2018adversarial,Zhang2019tradeoff,cui2020learnable,wang2021adaptive,Wang2020tradeoff,wu2020adversarial,Wang2021FA}. Kannan et al.~\cite{58kannan2018adversarial} proposed adversarial logit pairing (ALP) method to encourage the logits from original inputs and adversarial examples in the CNN model to be similar. Zhang et al.~\cite{Zhang2019tradeoff} provided an analysis of theoretically principled trade-off where the output from original inputs and adversarial examples are regularized by a KL-divergence function. Wang et al.~\cite{Wang2020Improving} explicitly differentiates the misclassified and correctly classified examples during the training and minimize the KL-divergence for misclassified examples. Cui et al.~\cite{cui2020learnable} proposed to parallelly train two models using original input and adversarial examples separately while the natural model is used to guide the training of robust model. Although the ideas behind these methods are intuitive for encouraging the same representation of original inputs and adversarial examples in CNN models, these regularizations arbitrary used point-to-point metrics and do not in practice align well with learning feature distributions. A cocurrent work is~\cite{levi2021domain}, they directly incorporate domain adaption into adversarial training without considering the class-level alignment. A fine-grained feature adaption scheme is still lacking.

\begin{figure*}[!t]
\centering
\includegraphics[width=0.9\linewidth]{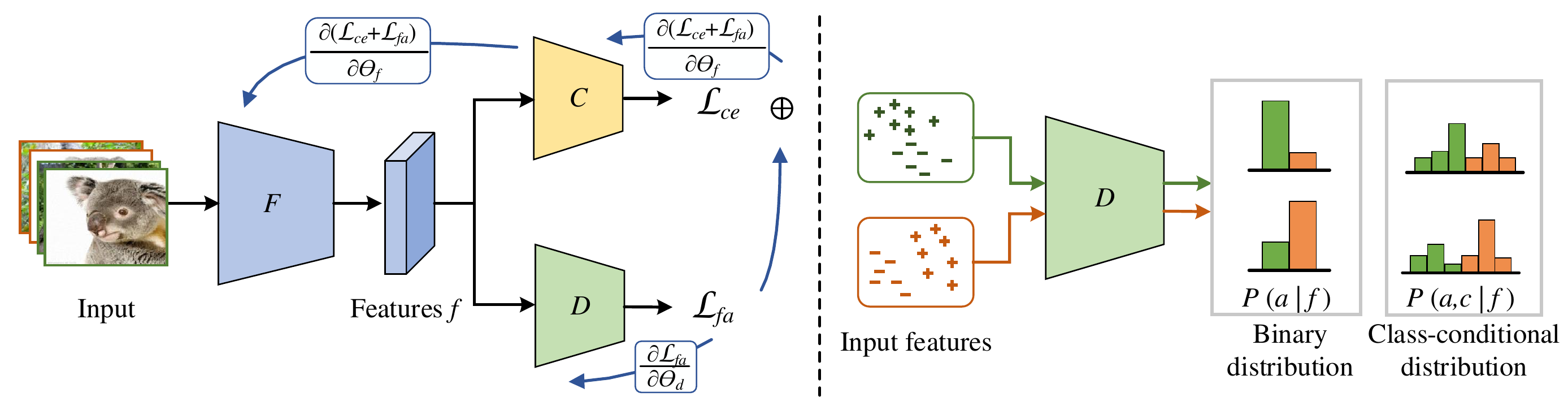}
    \caption{Overview of the proposed FAAT. \textbf{Left}: The training process of FAAT. We adopt a discriminator $D$ in the adversarial training framework to distinguish the feature distribution across natural data and adversarial examples, while the network $G$ is trained to generate invariant features against adversarial attacks. Here we divide $G$ into a feature extractor $F$ and a classifier $C$ for clear illustration, where $G=F \circ C$. \textbf{Right}: The class-conditional discriminator. We change the traditional discriminator which uses binary domain label into a class-conditional discriminator with multiple output to encourage the fine-grained class-level feature adaption.}
    \vspace{-15pt}
\label{fig:method}
\end{figure*}

\section{Methodology}
%In this section, we first conduct a study and find that the features generated from natural data and adversarial examples showing incremental deviations. Based on this observation, we design an domain adversarial loss term by a domain discriminator to encourage the similarity between natural and adversarial features.
Adversarial training aims at improving the model's robustness against adversarial attacks by training model on online-generated adversarial examples. In this paper, we focus on maintaining the trade-off between natural and robust accuracy in adversarial training by alleviating the shift between feature distribution of natural data and adversarial examples. Formally, given a CNN model $G$ trained on dataset $\mathcal{X}$ for $K$-class classification, We denote a natural data pair $(x,y)\in \mathcal{X}$ where $y\in \{0,...,K-1\}$. Adversarial examples are generated by adding a perturbation over $x$, i.e., $x^{adv}=x+\delta$. The goal is to learn a model $G$ which could achieve a low expected risk on both $x$ and $x^{adv}$.   

We discuss our method in the context of feature generation in CNN models. For more clear illustration, we divide the model $G$ into a feature extractor $F$ and a multi-class classifier $C$ , where $G=F \circ C$. In the following, we first introduce the class-conditional feature adaption scheme and then discuss the objective and the training procedure of our proposed FAAT framework.

\subsection{Class-conditional feature adaption}
Recent success in domain adaption~\cite{Haoran_2020_ECCV} reveal that feature distribution captured by a discriminator can be used to better measure the feature discrepancy between two domains. In these frameworks, a discriminator is usually equipped to model distribution $P(a\mid f)\in [0,1]$ given feature $f$, where $a$ denotes the data domain. By learning features that can confuse the discriminator, i.e., expecting $P(a=0\mid f) \approx P(a=1\mid f)$ where 0 and 1 stands for different domains, the features can be domain-invariant. Inspired by this, we first propose to adopt such a discriminator to encourage the invariant feature generation against the change of adversarial attacks. Specifically, the discriminator is trained to model distribution with the objective of $P(a=0\mid f)=1$ and $P(a=1\mid f^{adv})=1$, where $a=0$ and $1$ denotes natural data and adversarial example attribute
%, $f$ and $f^{adv}$ denotes the features of natural data and adversarial examples. 
Then the CNN model can be trained to fool the discriminator, i,e, expecting $P(a=0\mid f^{adv}) = 1$. We expect $F$ in $G$ to generate the domain-invariant features by solving:
\begin{equation}
\begin{aligned}
\label{eq:shiftloss}
		\min _{F} \mathcal{L}_{fa}&=-\sum \limits_{i=1} ^{n}\log D( F(x_i+\delta_i))\\
		&=-\sum \limits_{i=1} ^{n}\log P(a=0\mid F(x_i+\delta_i)),
	  \end{aligned}
\end{equation}
where $\delta$ can be optimized with PGD~\cite{madry2018towards}, and subjected to $\|\delta_i\|_{\infty} \leq \epsilon$. By minimizing $\mathcal{L}_{fa}$, the features generated from adversarial examples are globally similar to the features from natural data. However, it is not precise to only focus on feature invariance but neglect the underlying class-level structure difference. To take class discrimination into account, we further expand the discriminator to model class conditional distribution according to~\cite{Haoran_2020_ECCV}. Specifically, the output of discriminator is changed into multiple channels as $P(a, c\mid f)\in [0,1]$, where $c\in \{0,...,K-1\}$ denotes the class label, and $P(a\mid f)=\sum_{c=1}^{K-1} P(a,c\mid f)$. With this design, the predicted probability for domains is represented as a probability over different classes in specific domain, which enables the class-discriminative but domain-invariant feature generation. Thus the equation in Eq.\,\eqref{eq:shiftloss} becomes:
\begin{equation}
\begin{aligned}
\label{eq:c-shiftloss}
		\min _{F}\mathcal{L}_{fa}&=-\sum \limits_{i=1} ^{n} \log D( F(x_i+\delta_i))\\
		&=-\sum \limits_{i=1} ^{n}\sum \limits_{k=1} ^{K}d_{i,k}\log P(a=0, c=k\mid F(x_i+\delta_i)),
\end{aligned}
\end{equation}
where $d_{i,k}$ represents the class knowledge, $d_{i,k}=1$ when $x_i$ belongs to the $k$ class, otherwise $d_{i,k}=0$. The expectation is to push $P(a=0,c=k\mid f^{adv}) = 1$, meaning that the feature distribution of adversarial examples belonging to $k$-th class are expected to be similar with natural data belonging to $k$-th class.

%It tries to give the probablity of feature $f$ belonging to $a$ domain. Motivated by this, we propose to adopt a discriminator to evaluate the feature discrepancy between natural data and adversarial examples. By employing such a discriminator, we can model the adversarial examples. The CNN model tries to confuse the discriminator, i.e., expecting $P(a=0\mid f^{adv})=1$ where $a=0$ denotes the natural data and $a=1$ stands for adversarial examples, the features become invariant to the change of adversarial attacks. Let us take a CNN model $G$ trained on dataset $\mathcal{X}$ for $K$-class classification as an example. We consider a natural data pair $(x,y)\in \mathcal{X}$ and $y\in \{0,...,K-1\}$. The goal is to learn a model $G$ which could achieve a low expected risk on both $x$ and $x+\delta$. 
%Generally, we divide the CNN model $G$ into a feature extractor $F$ and a multi-class classifier $C$, where $G=C\circ F$. 

\subsection{Adversarial training with feature adaption}

We propose to incorporate the class-conditional feature adaptive objective in Eq.\,\eqref{eq:c-shiftloss} into the adversarial training process to induce the trade-off between natural and robust accuracy. The overall training framework is illustrated in the left part of Figure\,\ref{fig:method}. Since our training method relies on a discriminator to align the features, we conduct the training process by alternatively optimizing the trained model and discriminator as follows:

\textbf{Step 1}: The discriminator $D$ is first trained to model the distribution of features from different domains. This can be achieved by:
\begin{equation}
\begin{aligned}
\label{eq:dloss}
		\min _{D}&\mathbb{E}_{(x) \sim p_{x,c}}\log D(F(x))\\
		&+\mathbb{E}_{(x^{adv}) \sim p_{x^{adv},c}}[1-\log D(F(x^{adv}))],
\end{aligned}
\end{equation}
where $D(F(x))=\log P(a, c\mid F(x))$. Note that $F$ is fixed during the optimization of $D$. The goal of discriminator $D$ is to distinguish the features of natural data and adversarial examples in class-level.

\textbf{Step 2}: The model $G$ is trained with the task loss on the adversarial examples and the feature adaption loss output from the class-conditional discriminator. The task loss is used to train discriminative features for classification. Here we use most popular classification loss, i.e., cross-entropy loss $\mathcal{L}_{ce}$ for classification. The feature adaption loss is used to push the model generate class-conditional invariant features that can fool the discriminator. The overall objective can be represented as:
\begin{equation}
\begin{aligned}
\label{eq:ob}
		\min _{G} \mathbb{E}_{(x, y) \sim \mathcal{X}}[ \mathcal{L}_{ce}(G( x+\delta), y)+\lambda\mathcal{L}_{fa}],
\end{aligned}
\end{equation}
where $\lambda$ is the weighting factor which is set as $0.015$ in our experiments. More adjustments to this factor are illustrated in Sec.\,\ref{sec:ablation}. 

The objective in Eq.\,\eqref{eq:ob} can be understood as adding a feature alignment constraint into adversarial training process. However, our method is fundamentally different from other trade-off methods. In our method, we employ a adversarial framework where a discriminator is equipped to model the feature distribution, which is more guaranteed than the MSE-based metric in ALP~\cite{58kannan2018adversarial} or KL distance-based metric in Trades~\cite{Zhang2019tradeoff} and MART~\cite{Wang2020Improving}. Moreover, our method should also be distinguished from that of~\cite{levi2021domain} and~\cite{qian2021improving}, which only focus on generate globally similar representations of natural data and adversarial examples, neglecting the class-level structure difference. In our method, we directly incorporate the class knowledge into the discriminator and ask for more fine-grained feature adaption. This procedure simultaneously minimizes the feature discrepancy between natural data and adversarial examples, and maximizes the discrimination among classes. The entire training process of FAAT is elaborated in Sec.\,\ref{sec:training}.

\subsection{Training routine}\label{sec:training}
Algorithm\,\ref{trianing} shows the pseudo-code of our proposed FAAT training routine. We first randomly initialize the weights of discriminator $D$ and the target CNN model $G$. Then for each batch of training data during the training phase, a PGD optimization is first applied on training data to generate the corresponding adversarial examples. To distinguish the feature distribution of natural data and adversarial examples, the discriminator is first updated based on Eq.\,\eqref{eq:dloss}. Next, we use the updated discriminator to compute the feature adaptive loss based on Eq.\,\eqref{eq:c-shiftloss} to encourage the generation of class-conditional invariant features. The model $G$ is then updated based on the feature adaptive loss combined with original task loss, as described in Eq.\,\eqref{eq:ob}. This training routine can effectively minimize the discrepancy between natural data and adversarial examples while maximizing the class-level discrepancy. Consequently, the overall robustness and natural accuracy can be improved. Note that during the inference phase, the discriminator component can be removed, and the trained model can be robust to adversarial examples.

	\begin{algorithm}[!t]
		\SetKwFunction{isOddNumber}{isOddNumber}
		\SetKwInOut{KwIn}{Input}
		\SetKwInOut{KwOut}{Output}
		\KwIn{
			Training data $\mathcal{D}=\{X, Y\}$ , perturbation bound $\epsilon$, training epoch $N$, batch size $B$, learning rate $lr$. 
		}
		\KwOut{Trained model $G$, $D$ with parameter $\theta_g$ and $\theta_d$.}
		Initialize model $G$ and $D$ randomly or with pre-trained configuration. \\
		Initialize $\delta$ from uniform $(-\epsilon, \epsilon)$.  \\
		\For{epoch = 1 ... $N$ }{
			\For{i = 1 ... $B$ }{
				%// Perform single-step adversarial attack on batch $i$ samples. \\
				$\delta_{i} \leftarrow \delta_{i} + \alpha \cdot \operatorname{sign}(\nabla_{\delta_{i}}; \mathcal{L}(G(x_{i}+\delta_{i}), y_{i}))$; \\
				$\delta_{i} \leftarrow \max(\min(\delta_{i}, \epsilon), -\epsilon))$;
				
				$x_{i}^{adv} \leftarrow x_{i} + \delta_{i}$; \\
				
				Compute $\mathcal{L}_d$ based on Eq.\,\eqref{eq:dloss};
				
				Update $D$: $\theta_d \leftarrow \theta_d - lr\nabla_{\theta_d} \frac{\partial \mathcal{L}_d}{\partial \theta_d}$;
				
				Compute $\mathcal{L}_{ce}$ and $\mathcal{L}_{fa}$ based on Eq.\,\eqref{eq:c-shiftloss} and\,\eqref{eq:ob} ;
				
				Update $G$: $\theta_g \leftarrow \theta_g -lr \nabla_{\theta_g} \frac{\partial (\mathcal{L}_{ce}+\mathcal{L}_{fa})}{\partial \theta_g} $;

			}
		}
		\KwRet{model $G$.}
		\caption{Feature Adaptive Adversarial Training (FAAT)}
		\label{trianing}
		
	\end{algorithm}
\section{Experiments}
In this section, we verify the effectiveness of our method on classic image classification datasets: CIFAR-10 and CIFAR-100. We use WideResNet-34-10 as the trained CNN model architecture throughout all our experiments. Due to the page limits, we show crucial experimental results in our paper and more experimental results are presented in the supplementary material.

\begin{table*}[!t]
    \centering
    \fontsize{9.5}{11}\selectfont
    \caption{Robustness of models trained with different adversarial training methods on CIFAR-10 and CIFAR-100 datasets. All statistics are evaluated against PGD/CW attacks with 20/50 iterations and a random restart for $\epsilon=8/255$. %We highlight the best result in \textcolor{blue}{blue} color, and the second in \textcolor{red}{red} color.
    }
    \label{tab:white-box-result}
        \begin{tabular}{c|l|c|ccccc}
        \hlinew{1pt}
        {Dataset} &
        {Defense method} &
         
          {{Natural}} &
          {{PGD-20}} &
          {{PGD-50}} &
         {{CW-20}} &
          {{CW-50}} &
          {AA} \\ \hlinew{1pt}
          
        \multirow{7}{*}{CIFAR-10}  
                  & Standard~\cite{madry2018towards}       & 85.23 & 48.93 & 48.63 & 48.74 & 48.28  &32.36 \\
                 % & +FREE (m=8)~\cite{shafahi2019adversarial}  & 85.75 & 45.76 & 45.52 & 44.95 & 44.45  &24.22 & 128.16\\
                 % & ATTA-1~\cite{zheng2020efficient}   & \XSolidBrush    & 83.36 & 50.05 & 49.90 & 49.02 & 48.75  & \textbf{\textcolor{blue}{263.34}} \\
                  & +ATTA~\cite{zheng2020efficient}     & 84.43 & 54.65 & 53.74 & 54.25 & 54.01  &50.09 \\
                  & +CCG~\cite{tack2021consistency}     &  {88.32} & 54.71  & \textbf{54.53}  & 54.34  & 54.12 &51.00 \\
                  %& +MART~\cite{Wang2020Improving}      & 85.75 & 49.31 & 49.08 & 48.16  & 48.04 &44.96  \\
                  & +ALP~\cite{58kannan2018adversarial} & 86.45 & 53.12 & 52.59 & 52.85  & 52.52 &44.96\\
                  
                  & +Trades($\lambda=6$)~\cite{Zhang2019tradeoff}      & 85.05 & 51.20 & 50.82 & 50.81 & 50.65  & 46.17\\
                  
                  %& +AWP~\cite{wu2020adversarial}    & \Checkmark  & 85.26  & \textcolor{blue}{57.78} &  57.66 & 56.05   & 56.00  & \\
                  
                  & +\textbf{Ours(Binary $D$)}  &\textbf{89.75} &54.31 &54.10 &52.52 &52.02  &49.85 \\
                  & +\textbf{Ours(Class-conditional $D$)}  & 86.51 & \textbf{55.18} & 54.32 & \textbf{54.79} &\textbf{54.46}&\textbf{52.13}\\\hline \hline
        
        \multirow{7}{*}{CIFAR-100} 
                  & PGD-10 ~\cite{madry2018towards}      & 60.29 & 26.84 & 26.44 & 26.44 & 26.25  &12.60 \\
           % & +FREE (m=8)~\cite{wu2020adversarial}         & 60.11 & 26.79 & 22.66 & 25.69 & 25.60  & \\
                 % & ATTA-1   & \XSolidBrush    & 59.07 & 21.58 & 22.82 & 21.14 & 20.92  &  \\
                  & +ATTA~\cite{zheng2020efficient}     & 55.09 & 23.23 & 23.00 & 22.85 & 22.73  &20.11  \\
                  & +CCG~\cite{tack2021consistency}     & 60.74  & 27.27  & 27.03  & 26.04 & 26.01 &23.22 \\

                   %& +MART~\cite{Wang2020Improving}     &61.72 &29.94  &29.62 &28.71  &28.65  &25.60 \\
                   
                   & +ALP~\cite{58kannan2018adversarial}  & 59.65 & 28.28 & 28.02 & 27.14 & 27.02  &22.60 \\
                  
                  & +Trades($\lambda=6$)~\cite{Zhang2019tradeoff}     & 62.37 & 25.31 & 25.02 & 24.53 & 24.21  & 22.24 \\
                  
                  %& AWP~\cite{wu2020adversarial}    & \Checkmark  & 59.11   & 33.62  & 33.6  & 31.01  & 30.98  & \\
                  
                  & \textbf{Ours(Binary $D$)}  &60.77 &27.87 &27.25 &26.45 &26.01 &23.51  \\
                  &  +\textbf{Ours(Class-conditional $D$)}  & \textbf{62.58} & \textbf{30.22} & \textbf{30.17} & \textbf{30.58} & \textbf{30.36}  & \textbf{27.87}   \\ 
                 \hlinew{1pt}
        \end{tabular}
        
    \end{table*}

%\subsubsection{Compared methods.} We compare the performance of our proposed SPGAT with well-known state-of-the-arts methods including standard PGD-10 adversarial training; FREE $m=8$~\cite{shafahi2019adversarial}, which uses single-step FGSM with 8 hop steps; ATTA-$k$: most related work that $k$-step iteration PGD adversarial training with transferable adversarial examples \cite{zheng2020efficient}. Here, we set $k=1$ and $10$ in ATTA for fair comparison. All the compared methods are performed using the default codes and settings as the authors released.

\subsection{Setup}
\noindent\textbf{Compared methods.} We compare the performance of our method with various counterparts, including standard adversarial training method~\cite{madry2018towards} which uses PGD-10 to generate on-line adversarial examples; methods focusing on maximizing the strength of adversarial examples for improving robust accuracy: ATTA~\cite{zheng2020efficient} and CCG~\cite{tack2021consistency}; and methods focusing on achieving trade-off between natural accuracy and robust accuracy:  ALP~\cite{58kannan2018adversarial} and Trades~\cite{Zhang2019tradeoff}. %HAT~\cite{rade2021helperbased}, and AWP~\cite{wu2020adversarial}.
\vspace{5pt}

\noindent\textbf{Training details.}
We implement all our experiments using PyTorch on an Intel Core i9 with 32GB of memory and an NVIDIA TITAN XP GPU. In our experiments, we alternatively update the discriminator and CNN model in each epoch. For the hyper-parameters setting, we generally follow the settings suggested by previous works~\cite{cui2020learnable,Zhang2019tradeoff}. Specifically, we use PGD-10 algorithm to generate adversarial examples in each training epoch, where perturbation constraint $\epsilon=8/255$, step size is set as $2\epsilon/k$. We train the discriminator and CNN model for 100 epochs using stochastic gradient descent (SGD) optimizer with an initial learning rate as $0.1$ and it is decay by a factor of $0.1$ at 50\% and 75\% of the total epoch. 

 \begin{table}[!t]
    \caption{Natural and robust accuracy (\%) of WideResNet34-10 models trained on CIFAR-100 dataset against black-box transfer attack.} %We choose $l_{\infty}$ threat model with $\epsilon=8/255$ for PGD. Specially, for CW$_2$, $\epsilon$ is fixed to $160/255$.s: adversarial examples are crafted from PGD-10 AT pre-trained PreActResNet18
    
    \label{transfer_attack}
    \centering
    %\resizebox{0.6\textwidth}{!}{
        \begin{tabular}{lcccc}
        \hlinew{1pt}
         {{Method}} 
 &Natural            & PGD-50         & CW$_{\infty}$   \\ \hlinew{1pt}

          Standard   & 60.29         & 42.13          & 56.55\\
         % Free   & 55.09       & 40.81          & 56.51\\
           +ATTA~\cite{zheng2020efficient}   & 55.09         & 39.79          & 50.74\\
           +CCG~\cite{tack2021consistency}    & 60.74          & \textbf{46.74}          & 58.92\\ 
          +ALP~\cite{58kannan2018adversarial}    &59.75          & 45.79          & 56.94 \\ 
            +Trades($\lambda=6$)~\cite{Zhang2019tradeoff}  & 62.37          & 42.09          & 53.86\\
           %+AWP ~\cite{wu2020adversarial} & 59.11          & 43.62          & 55.97 \\ 
           %\rowcolor{mygray} 
            +\textbf{Ours}   & \textbf{62.58}        & 44.65          & \textbf{59.47}
          \\ \hlinew{1pt}
        \end{tabular}
    %}
    \end{table}

\subsection{Main results}

%\subsubsection{White-box attacks}
\noindent\textbf{White-box attacks.} We first evaluate the robustness of our proposed method under white-box attacks where the adversarial examples are generated by the known model. Here we use three attack methodologies: Common attacks including PGD-$k$ and CW-$k$, and stronger attack AA.
%PGD with 20 and 50 inner iterations on both cross-entropy loss (PGD-$k$) and the Carlini-Wagner loss (CW-$k$) (step size is set as $2\epsilon/k$, where $k$ is the iterations number). We also generate adversarial examples with more strong attack, Auto-Attack (AA), which is known as the strongest attack by far. 
%The training time for each epoch is also tested for evaluating the efficiency of each method. The results are reported in Table\,\ref{tab:white-box-result}. 
As we can see in Table\,\ref{tab:white-box-result}, focusing on strengthening the adversarial attack in ATTA can significantly improve the robustness accuracy but also lead to a decrease in natural accuracy on CIFAR-10 dataset. CCG uses specific designed data augmentation to increase the data diversity so it achieves higher natural and robust accuracy in CIFAR-10 dataset. However, both the ATTA and CCG methods degrade in CIFAR-100 dataset. 
%It makes us wonder whether the excessive data enhancement or augmentation will induce overfitting in datasets with more classes.
For the methods designed for trade-off terms, adding Trades or ALP into the standard adversarial training can somewhat maintain the trade-off between natural accuracy and robust accuracy, but neither of them can achieve overall improvements over the standard training. Instead, our method focus on class-conditional feature adaption across natural data and adversarial examples, and can effectively enhance both natural and robustness accuracy, especially for CIFAR-100 dataset with more classes, as the natural and robust accuracy against AA achieves $62.58\%$ and $27.87\%$, respectively, surpassing baselines with a large margin.

%It can be observed that, compared with the state-of-the-art methods, the proposed SPGAT could effectively improve the robustness against $l_{\infty}$ and $l_{2}$ adversaries on both CIFAR-10 and CIFAR-100 datasets. For $l_{1}$ adversaries, ATTA-1 achieves best result on CIFAR-10 dataset but degrades dramatically on large-scale CIFAR-100 dataset. Similarly, FREE ($m=8$) has the best result on CIFAR-10 dataset but not on CIFAR-10 dataset. This reveals the instability and limitations of these methods. On contrarily, our proposed SPGAT can have stable and comparable performance on both datasets.

%It is worth noting that our method is especially effective against $l_{\infty}$ and $l_{2}$ adversaries compared to the baselines, but it is more difficult to achieve robustness against $l_{1}$ adversaries.

    \begin{table}[]
    \centering
    %	\vspace{-8pt}
    \caption{Natural and adversarial robust accuracy (\%) of the WideResNet34-10 models trained by combinations of adversarial training methods.}
    \label{tab:combination}
    
   \fontsize{9}{11}\selectfont
    \begin{tabular}{lcccc}
    \hlinew{1pt}
         \multirow{2}{*}{Method}    & \multicolumn{2}{c}{CIFAR-10} & \multicolumn{2}{c}{CIFAR-100}\\ \cline{2-5}
         &Natural  & {PGD-20} &Natural & {PGD-20}   \\ \hlinew{1pt}
         %Free  & 85.75 & 45.76  & 60.11 & 26.79\\
        %\rowcolor{mygray}  
         %Free+Ours &  &  & & \\ 
       ATTA & 84.43 & 54.65  & 55.09 & 23.23  \\
       \rowcolor{mygray} 
        ATTA+Ours & \textbf{87.33}  & 53.84 & \textbf{59.01} & \textbf{25.32}  \\
        
         Trades($\lambda=6$)  & 85.05 & 51.20  & 62.37 & 25.31   \\
          \rowcolor{mygray} 
        Trades+Ours  & 83.22  &\textbf{54.96}  &60.51   &\textbf{28.18}   \\
        ALP  &86.45   &53.12  &59.65 & 28.28  \\
        \rowcolor{mygray} 
        ALP+Ours  &84.20  &\textbf{56.33}  &\textbf{61.35}  &\textbf{29.58}   \\
        CCG  & 88.32 & 54.71   & 60.74 & 27.27 \\
        \rowcolor{mygray} 
        CCG+Ours  &87.31  &\textbf{55.14}  &\textbf{63.69}  &\textbf{30.95}  \\

         \hlinew{1pt}
    \end{tabular}%
    %}
\end{table}

%\subsubsection{Black-box transfer attacks}
\vspace{5pt}
\noindent\textbf{Black-box transfer attacks.} To evaluate the effectiveness of our proposed FAAT in more practical defense scenarios, we test the model under black-box transfer attacks, where adversarial examples are generated from a source model and then transfer to the target model. In this experiment, adversarial examples are crafted from PreActResNet18 (source model) trained with standard adversarial training (PGD-10), and we use PGD-50 and CW$_{\infty}$ as black-box adversaries. For the target model, we also use WideResNet-34-10 as the model architecture. The results are shown in Table\,\ref{transfer_attack}. It is obvious that the black-box attacks are weaker than white-box attacks, as the robust accuracy denoted in Table\,\ref{transfer_attack} are double times higher than that against white-box attacks in Table\,\ref{tab:white-box-result} on the same dataset. From the comparison with other methods, we can see that our method can achieve comparable results with CCG which applied data augmentation in their method. In addition, our method shows consistent improvement in natural accuracy by taking trade-off between natural and robust accuracy into consideration.

\vspace{5pt}
%\subsubsection{Combing with other methods}
\noindent\textbf{Combing with other methods.} Our FAAT provides an universal framework for adversarial training. To verify the flexibility of our method, we combine various methods with our FAAT framework. The evaluation is under the white-box attack following the same setting in Sec.\,\ref{sec:training}. The results are summarized in Table\,\ref{tab:combination}. On CIFAR-10 dataset, using our FAAT framework can effectively improve the existing adversarial methods on the trade-off between natural and robust accuracy. For ATTA which focuses on enhancing attack strength, equipped with our FAAT framework, the ATTA method can improve the natural accuracy by $3\%$. For Trades and ALP, using the FAAT framework can improve the robustness accuracy while getting rid of a dramatic decrease in natural accuracy. On CIFAR-100 dataset, FAAT framework can help the existing methods improve the overall performance by incorporating the class-conditional feature adaption into the training process. Particularly, our FAAT framework improves CCG by $3\%$ in both natural and robust accuracy, which creates new SOTA on CIFAR-100 dataset.

 \begin{table}[!t]
    \caption{Training time (s) in each epoch for different adversarial training methods.} %We choose $l_{\infty}$ threat model with $\epsilon=8/255$ for PGD. Specially, for CW$_2$, $\epsilon$ is fixed to $160/255$.
    \label{tab:trades}
    \centering
    \resizebox{0.48\textwidth}{!}{
        \begin{tabular}{lcccccc}
        \hlinew{1pt}
         {{Dataset}} 
 &Standard            & ATTA~\cite{zheng2020efficient}         & CCG~\cite{tack2021consistency} & ALP~\cite{58kannan2018adversarial} & Trades~\cite{Zhang2019tradeoff} & Ours    \\ \hlinew{1pt}

          CIFAR-10   & 1235.10         & 1425.79          & 3556.23 &2033.20 &2199.24 &1440.12\\
         % Free   & 55.09       & 40.81          & 56.51\\
            CIFAR-100   & 1234.11         & 1429.49          & 3567.37 &2042.35 &2205.41 &1445.24\\
 \hlinew{1pt}
        \end{tabular}
    }
    \end{table}   
    
\begin{figure}[!t]
\centering
\includegraphics[width=1.0\linewidth]{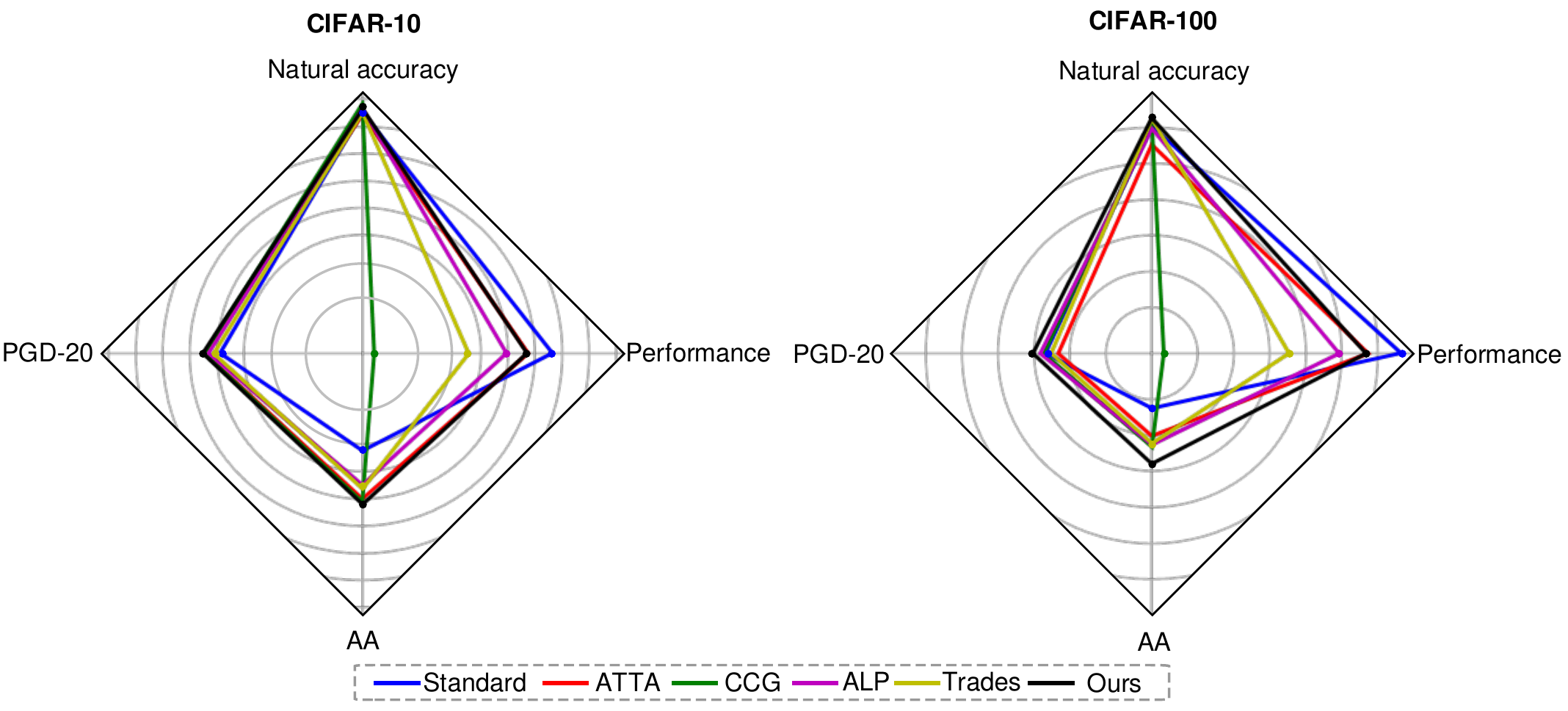}
    \caption{Illustration of natural accuracy, robust accuracy against PGD-20, robust accuracy against AA, and performance trade-offs using different adversarial training methods.}
\label{fig:trade}
\end{figure}

\subsection{Efficiency analysis}
A bottleneck of adversarial training is the expensive training time due to the on-line data augmentation process. In this part, we analyze the efficiency of our proposed FAAT. Compared to standard adversarial training, our FAAT additionally equips a discriminator. Since we use the feature extractor in the target model to extract features thus the discriminator architecture is simple, and training such a discriminator only takes trivial additional consumption cost. We report the training time for each epoch of different methods in Table\,\ref{tab:trades}. As we can see, the training of FAAT is faster than Trades or ALP, and much faster than CCG which needs to deliberately augment additional training data. 

To better understand the trade-offs among natural accuracy, robust accuracy, and efficiency,
Figure\,\ref{fig:trade} demonstrates a four-dimensional radar plot with natural accuracy, robust accuracy against PGD-20, robust accuracy against AA, and performance on four axes. Note that we plot the negative of the running time for each training epoch on the performance axis. Thus, the ideal adversarial training method exhibits highly rhombus-shape quadrangle. We find that our FAAT (denoted by the black solid quadrangle) gives a better trade-off on efficiency and effectiveness on adversarial training. Hence, we believe the FAAT can provide an universal and efficient framework for adversarial training while free of burden.

\subsection{Feature distribution}
To verify whether incorporating such a discriminator in adversarial training can improve the feature learning of CNN models, we design an experiment to investigate the feature distribution before and after our FAAT. We first intuitively show the feature distribution of natural data and adversarial examples using t-SNE embedding~\cite{kang2019contrastive} on CIFAR-10 dataset in Figure\,\ref{fig:distribution}. Here we generate the adversarial examples by PGD-20 attack, and the features are extracted from the last convolution layer before normalization. Compared to the baseline training, as expected, the data distribution learned by FAAT demonstrates higher intra-class compactness and much larger inter-class margin. Moreover, the features from natural data and adversarial examples have similar distributions, as can be seen in Figure\,\ref{fig:distribution}\,(c) and Figure\,\ref{fig:distribution}\,(d). This suggests that our FAAT can generate more class-discriminative but invariant features against adversarial attacks, which is consistent with the results in Table\,\ref{tab:white-box-result}. %Please see the appendix for more intuitive feature distribution comparisons.

\begin{figure*}[!t]
\centering
\includegraphics[width=0.9\linewidth]{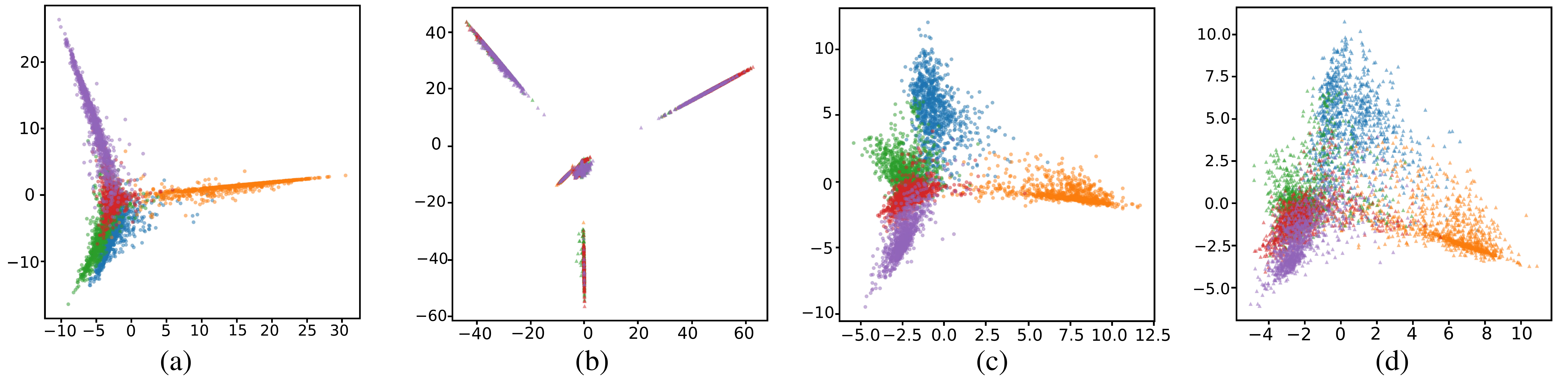}
    \caption{Visualization of feature distribution in WideResNet-34-10 model trained wo/w our proposed FAAT on CIFAR-10 dataset. (a)-(b) Feature distribution of natural data and adversarial examples in model trained with baseline method. (c)-(d) Feature distribution of natural data and adversarial examples in model trained with FAAT method.}
\label{fig:distribution}
\end{figure*}

\begin{figure}[!t]
\centering
\includegraphics[width=0.9\linewidth]{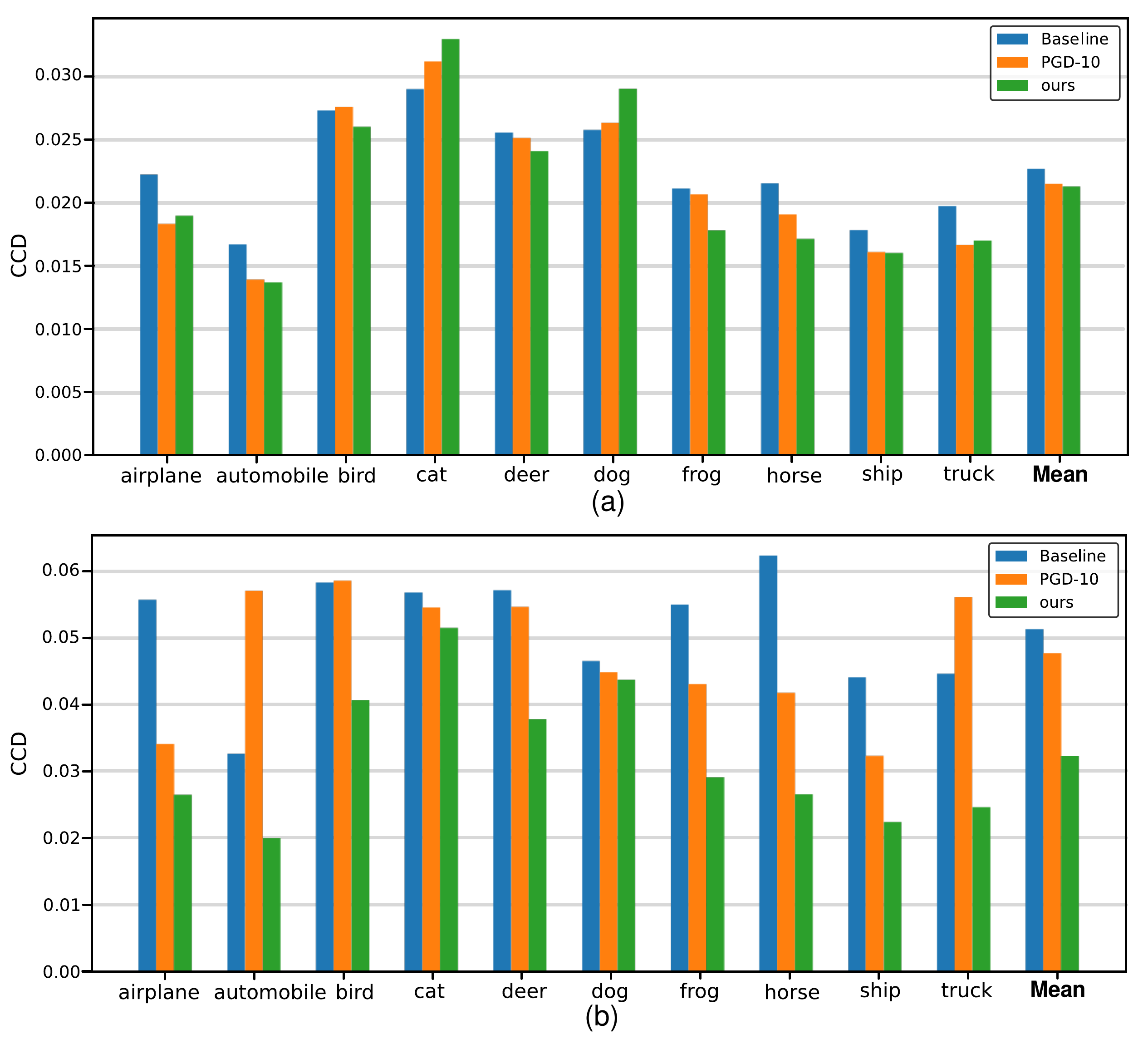}
    \caption{CCD analysis of the feature distributions produced by models trained on CIFAR-10 dataset. We show the CCD value using Eq.\,\eqref{eq:ccd} for each class and overall mean value on (a) natural data and (b) adversarial examples.}
\label{fig:ccd}
\end{figure}

Moreover, we use the Class Center Distance (CCD) suggested in~\cite{Haoran_2020_ECCV} as a metric to evaluate the feature distribution by computing intra-class compactness over inter-class distance. The CCD for class $k$ can be represented as:
\begin{equation}
\begin{aligned}
\label{eq:ccd}
		CCD(k)=\frac{1}{K-1}\sum \limits_{i=1,i\neq k} ^{K}\frac{\frac{1}{\mid S_k\mid}\sum_{\mathbf{f}\in S_k}\|\mathbf{f}-\mu_i\|^2}{\|\mu_k-\mu_i\|^2},
\end{aligned}
\end{equation}
where $\mu_k$ is the class center of class $k$, and $S_k$ denotes the set of all features belonging to class $k$. Note that a lower CCD value suggests a more densely clustering. The results are provided in Figure\,\ref{fig:ccd}. Our FAAT achieves a much lower CCD on most classes and the lowest mean CCD value on both natural data and adversarial examples. It indicates that FAAT can achieve better class-level feature adaption across natural data and adversarial examples.

\subsection{Ablation studies}\label{sec:ablation}
\noindent\textbf{Effect of class knowledge encoding.} We compare FAAT using the class-conditional discriminator with that uses the traditional binary discriminator, to verify the merits of introducing class-level knowledge in adversarial training. The results are shown in the last two rows in Table\,\ref{tab:white-box-result}. It is obvious that incorporating the class-level knowledge into feature adaption can effectively improves the adversarial training performance, especially for CIFAR-100 dataset with more classes. We believe the reason is that the incorporation of class-level clustering helps improve the overall performance.
\vspace{5pt}

\begin{figure}[!t]
\centering
\includegraphics[width=0.9\linewidth]{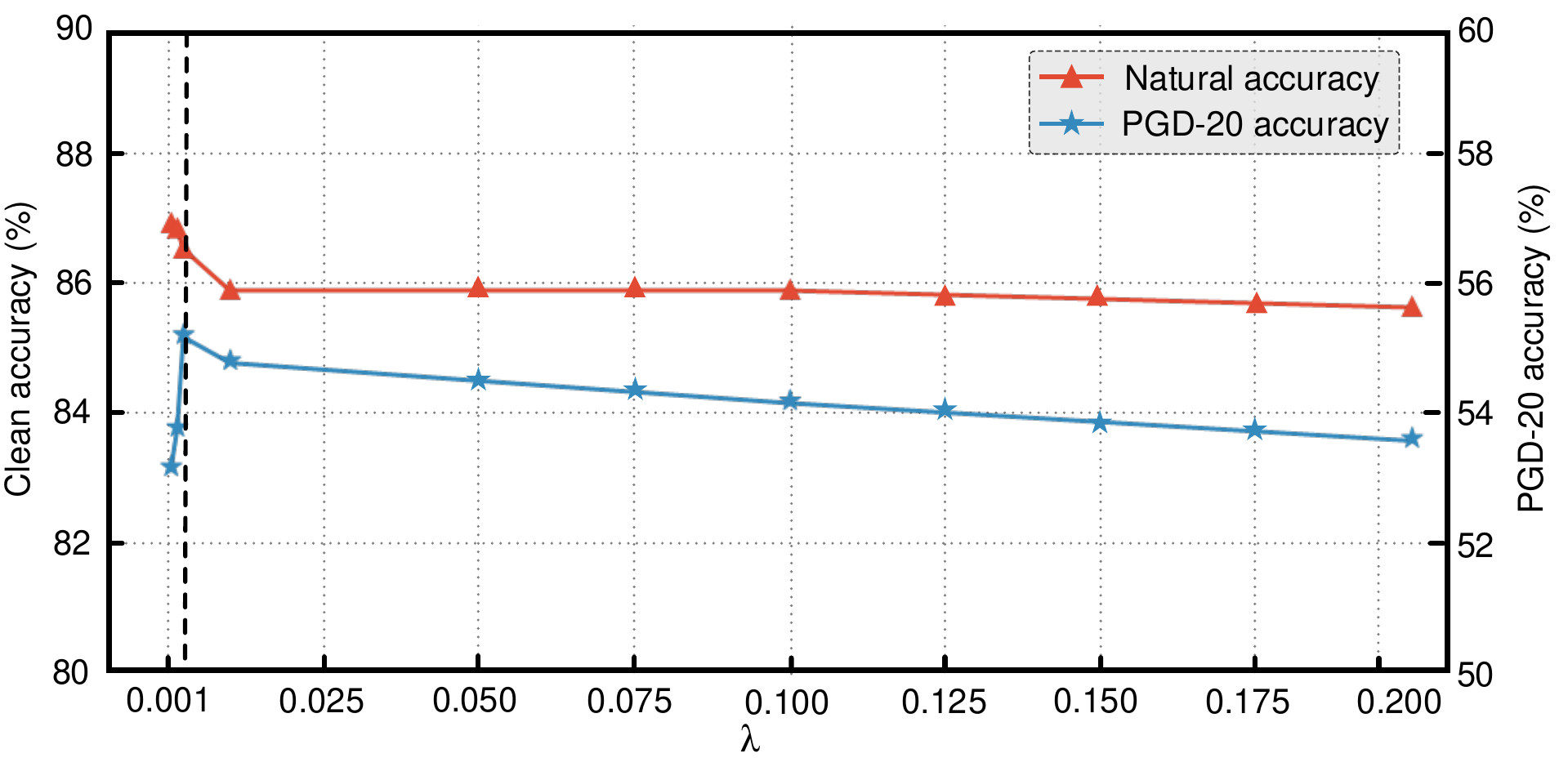}
    \caption{The curve of natural accuracy and robust accuracy of WideResNet-34-10 trained on CIFAR-10 dataset using FAAT with different values of $\lambda$.}
\label{fig:lamada}
\end{figure}

\noindent\textbf{Hyper-parameter sensitivity}. We study the sensitivity of FAAT to the balance weight $\lambda$ on CIFAR-10 dataset. Generally, we follow the setting in~\cite{Haoran_2020_ECCV} to initialize $\lambda=0.001$ and test the values floating up or down. The results are provided in Figure\,\ref{fig:lamada}. 
%In a vast range, the performance of FAAT outperforms the . 
As $\lambda$ gets larger, the overall accuracy steadily increases before decreasing. We set $\lambda=0.0015$ throughout the experiments as the setting denotes the best performance.

\section{Conclusion}\label{sec:conclusion}
In this paper, we focus on maintaining the trade-off between natural accuracy and robust accuracy in adversarial training by proposing a class-conditional feature adaptive adversarial training framework. A class-conditional discriminator is incorporated in adversarial training to guide the feature adaption across natural data and adversarial examples. Comprehensive experiments and analysis validate the effectiveness of our FAAT, where our method achieves the best robust accuracy in AA while maintaining the natural accuracy.

%%%%%%%%% REFERENCES
{\small
\bibliographystyle{ieee_fullname}
\bibliography{egbib}
}

\newpage
\onecolumn
%%%%%%%%% TITLE - PLEASE UPDATE

\title{\centering \Large \textbf{Supplementary Material}\\ 
\begin{center}
    \large \textbf{Push Stricter to Decide Better: A Class-Conditional Feature Adaptive Framework for Improving Adversarial Robustness}
\end{center} }  % **** Enter the paper title here

\maketitle
\thispagestyle{empty}
\appendix

%%%%%%%%% BODY TEXT - ENTER YOUR RESPONSE BELOW
\section{Implementation Details}
\subsection{Discriminator architecture}
\noindent\textbf{Network architecture.} In our FAAT framework, the discriminator is used to distinguish the feature distribution belongings in a class level. Since the input of the discriminator is the features generated by the target trained model, we can design the discriminator with a simple architecture. We adopt the discriminator architecture in~\cite{Haoran_2020_ECCV}, which comprises two $3\times 3$ convolutional layers followed by LeakyReLU activation function, and then two fully connected layers are applied to predict the class-conditional possibility over natural and adversarial domain, respectively.
\vspace{5pt}

\noindent\textbf{Class knowledge encoding.} The labels for training a traditional discriminator are binary, namely [1,0] and [0,1], for the natural and adversarial data domain respectively. To incorporate the class knowledge into the discriminator, the labels are transferred into $[\textbf{d};\textbf{0}]$ and $[\textbf{0};\textbf{d}]$, where $\textbf{d}\in \mathbb{R}^{K\times 1}$ denotes the class knowledge, and $\textbf{0}\in \mathbb{R}^{K\times 1}$ is an all-zero vector. In our task, the labels of adversarial examples are known so we can directly use one-hot hard labels for generating $\textbf{d}$. 

\subsection{Combination with other methods}
FAAT provides an universal framework for adversarial training. It does not conflict with other adversarial training methods. For the combination of existing methods with FAAT framework, we directly add the feature adaption loss in Trades and ALP. For ATTA and CCG, we describe the details as follows.

\vspace{5pt}
\noindent\textbf{ATTA+Ours.} Zheng et al.~\cite{zheng2020efficient} showed that reusing the adversarial examples generated from last epoch as the starting point in the next epoch can effectively strengthen the adversarial attacks. This process can be presented as:
\begin{equation}
\begin{aligned}
\label{eq:ob}
		\min _{G} \mathbb{E}_{(x, y) \sim \mathcal{X}}[ \mathcal{L}_{ce}(G( x^{adv}_t), y)], ~~\text{s.t.}~~~ x^{adv}_t = x^{adv}_{t-1}+\delta,
\end{aligned}
\end{equation}
where $\delta$ denotes the perturbation and is optimized using PGD-10. But only focusing on strengthening the adversarial attacks can lead to a bigger shift between natural data and adversarial examples. As reported in their paper~\cite{zheng2020efficient}, the robust accuracy against PGD-20 achieves 10\% improvements over standard adversarial training but the natural accuracy drops by 5\%. By incorporating our FAAT framework, the training objective becomes:
\begin{equation}
\begin{aligned}
\label{eq:ob}
		\min_{G} \max_{D} \mathbb{E}_{(x, y) \sim \mathcal{X}}[ \mathcal{L}_{ce}(G( x^{adv}_t), y)-\lambda D(f( x^{adv}_t))],  ~~\text{s.t.}~~~ x^{adv}_t = x^{adv}_{t-1}+\delta,
\end{aligned}
\end{equation}
where $f( x+\delta)$ denotes the features in $G$ before Softmax function.
\vspace{5pt}

\noindent\textbf{CCG+Ours.} Tack et al.~\cite{tack2021consistency} proposed a consistency regularization to augment additional data for improving the robust generalization. Given two augmentations $T_1,T_2 \sim \mathcal{T}$, the objective of CCG can be represented as:
\begin{equation}
\begin{aligned}
\label{eq:ob}
		\min _{G} \mathbb{E}_{(x, y) \sim \mathcal{X}}[ \frac{1}{2}\sum_{i=1}^2\mathcal{L}_{ce}(G( T_i(x)+\delta), y)+\lambda \cdot \text{JS}(\hat{G}( T_1(x)+\delta_1;\tau)\|\hat{G}( T_2(x)+\delta_2;\tau))],
\end{aligned}
\end{equation}
where $\text{JS}(\cdot \| \cdot)$ denotes the Jensen-Shannon divergence, $\hat{G}(x; \tau)$ represents a temperature-scaled distribution, and $\hat{G}(x; \tau)=\text{Softmax}(f(x)/\tau)$. The motivation of CCG is to accomplish the consistency over data augmentations. To combine with FAAT, we have
\begin{equation}
\begin{aligned}
\label{eq:ob}
		\min _{G}\max _{D} \mathbb{E}_{(x, y) \sim \mathcal{X}}[ \frac{1}{2}\sum_{i=1}^2\mathcal{L}_{ce}(G(T_i(x)+\delta), y)+\lambda_{1} \cdot \text{JS}(\hat{G}( T_1(x)+\delta_1;\tau)\|\hat{G}( T_2(x)+\delta_2;\tau)) \\
		-\lambda_{2}\cdot \frac{1}{2}\sum_{i=1}^2 D(f(T_i(x)+\delta))].
\end{aligned}
\end{equation}

\section{Additional Experiments}
\subsection{More experiments on CIFAR-10 and CIFAR-100 datasets} 
\noindent\textbf{Unseen adversaries.} We consider a wide range of unforeseen adversaries, e.g., robustness on different attack threat radii $\epsilon$, or even on different norm constraints (e.g., $l_{2}$ and $l_{1}$), in order to further measure the robustness of trained models against multiple perturbation. The results on CIFAR-10 and CIFAR-100 are reported in Table\,\ref{tab:unseen} and Table \ref{tab:unseencom}, respectively. 

For CIFAR-10 dataset, as shown in Table\,\ref{tab:unseen}, incorporating FAAT into standard adversarial training can remarkably improve the robustness against unseen attacks. Although ALP and Trades methods share the same motivation with us that encourages the similar distribution of natural and adversarial data, incorporating FAAT still brings improvements under most attack scenarios which is benefited from the fine-grained feature adaption. CCG improves the model robustness via augmenting additional data to increase the diversity of the dataset. As our FAAT encourages similar distribution across natural data and adversarial examples, it will weaken the augmentation strength, thus leading to a slight decrease under some attack conditions on CIFAR-10 dataset. Fortunately, we take the class knowledge into consideration in FAAT, which comes to play an important role when the the scale of dataset becomes larger. As we can observe in Table\,\ref{tab:unseencom}, with the class-conditional feature adaption scheme, our proposed FAAT can significantly help the existing methods improve the adversarial robustness against unseen attacks. Especially for CCG method, the CCG+Ours outperforms CCG under every attack scenario, which makes us believe the proposed FAAT has more potential for model robustness improvements on large-scale datasets which exist in more practical applications.

%We also test the combination of existing adversarial methods with our FAAT on CIFAR-100 dataset with unseen data, and report the results in Table\,\ref{tab：unseencom}. As we can see, equipping with the FAAT framework can effectively improve the overall performance of various adversrial training methods.

     \begin{table*}[!h]
        \centering
       % \fontsize{9}{11}\selectfont
        \caption{Natural and robust accuracy(\%) of WideResNet-34-10 trained with $l_{\infty}$ of $\epsilon=8/255$ boundary against unseen attacks on CIFAR-10 dataset. For unseen attacks, we use PGD-50 under different sized $l_\infty$ balls, and other types of norm ball, e.g., $l_{2}$, $l_1$.}
        \label{tab:unseen}
        \begin{tabular}{lcccccc}
        \hlinew{1pt}

         \multirow{2}{*}{{Method}} 
        & \multicolumn{2}{c}{$l_{\infty}$} & \multicolumn{2}{c}{$l_2$} & \multicolumn{2}{c}{$l_1$} \\ \cline{2-7} 
       & 4/255          & 16/255          & 150/255     & 300/255     & 2000/255       & 4000/255      \\ \hlinew{1pt}

          Standard  & 67.92          & 21.52           & 52.49       & 24.93       & 67.36          & 46.99 \\
           \rowcolor{mygray}
          +\textbf{Ours}  & \textbf{73.26}          &\textbf{ 22.68}           & \textbf{58.25 }      & \textbf{25.61 }      & \textbf{70.59 }         & \textbf{50.16} \\
  %& FREE (m=8)    &42.50          & 8.74      & 34.63      & 15.75       & 46.32          & 34.82  \\
      %  & ATTA-1  &35.18          & 13.28     & 26.77      & 15.73       & 39.10          & 26.32  
       
         ALP~\cite{Wang2020Improving}    & 72.76          & 22.98           & 56.96       & 24.70       & 69.53          & 48.77  \\  \rowcolor{mygray}
        +\textbf{Ours}     & 71.49 & \textbf{26.44} & \textbf{59.96} & \textbf{30.10}  & \textbf{69.56}    & \textbf{52.71}    \\
         Trades~\cite{Zhang2019tradeoff}  & 72.25          & 24.12           & 58.71       & 27.96       & 69.41          & 51.18 \\ 
          \rowcolor{mygray}
         +\textbf{Ours}     & 71.21 & \textbf{24.81} & \textbf{58.73} & \textbf{28.99 } & \textbf{69.51}    & \textbf{52.01}\\
           CCG\cite{tack2021consistency}   & 73.46          & 23.86           & 58.13       & 28.98       & 72.81          & 54.16 \\ 
            \rowcolor{mygray}
        +\textbf{Ours}  & \textbf{73.55} & 22.97 & \textbf{58.83} & 25.84  & 71.17    & 50.45 \\
        %& AWP     &46.24          & 14.27     & 38.53      & 19.72       & 47.61          & 36.10  \\ 
        \hlinew{1pt} 
        \end{tabular}
    \end{table*}

  \begin{table*}[!h]
        \centering
       % \fontsize{9}{11}\selectfont
        \caption{Natural and robust accuracy(\%) of WideResNet-34-10 trained with $l_{\infty}$ of $\epsilon=8/255$ boundary against unseen attacks on CIFAR-100 dataset. For unseen attacks, we use PGD-50 under different sized $l_\infty$ balls, and other types of norm ball, e.g., $l_{2}$, $l_1$.}
        \label{tab:unseencom}
        \begin{tabular}{lcccccc}
        \hlinew{1pt}

         \multirow{2}{*}{{Method}} 
        & \multicolumn{2}{c}{$l_{\infty}$} & \multicolumn{2}{c}{$l_2$} & \multicolumn{2}{c}{$l_1$} \\ \cline{2-7} 
      &  4/255          & 16/255          & 150/255     & 300/255     & 2000/255       & 4000/255      \\ \hlinew{1pt}
       % \multirow{10}{*}{CIFAR-10}  
       % & Standard     &85.23 & 67.92          & 21.52           & 52.49       & 24.93       & 67.36          & 46.99 \\
        % & +\textbf{Ours}  & \textbf{86.51}   & \textbf{73.26}          & \textbf{22.68}           & \textbf{58.25}       & \textbf{25.61}       & \textbf{70.59}          & \textbf{50.16} \\
        %& FREE (m=8) & 64.66          & 15.94           & 51.86       & 26.74       & 64.15          & 46.56 \\
       % & ATTA-1     & 67.99          & 16.95           & 58.85       & 28.50       & 72.47          & 57.97 \\
       % & ATTA~\cite{zheng2020efficient} & 84.43   & 69.94          & 22.43           & 58.15       & 28.09       & 70.91          & 53.89 \\
       %  & +\textbf{Ours}  \\
       % & CCG~\cite{tack2021consistency}     & \textbf{88.32}   & \textbf{73.46}          & 23.86           & \textbf{59.13}       & \textbf{28.98}       & \textbf{72.81}          & \textbf{54.16} \\ 
        % & +\textbf{Ours}   & 87.31  & 73.55 & 22.97 & 58.83 & 25.84  & 71.17    & 50.45 \\
       % & MART~\cite{Wang2020Improving}   & 83.62     & 72.76          & 22.98           & 56.96       & 24.70       & 69.93          & 48.77 \\ 
        % & +\textbf{Ours} \\
       % & Trades~\cite{Zhang2019tradeoff}  & 84.42   & 72.25          & \textbf{25.12}           & 58.91       & 27.96       & 69.91          & 51.18 \\ 
        % & +\textbf{Ours}   & 83.22  & 71.21 & 24.81 & 58.73 & 28.99  & 69.51    & 52.01 \\
        %& AWP        & 73.74          & 25.49           & 61.27       & 29.62       & 72.20          & 53.86 \\ \rowcolor{mygray} 
      %  \hline \hline
                    
        %\multirow{10}{*}{CIFAR-100} 
          Standard  &40.67          & 9.96      & 30.69      & 12.99       & 42.43          & 28.24  \\
                 \rowcolor{mygray}
          +\textbf{Ours}  &\textbf{46.01}          & \textbf{11.70  }   & \textbf{35.35}      & \textbf{15.55}       & \textbf{46.64}          & \textbf{31.87} \\
  %& FREE (m=8)    &42.50          & 8.74      & 34.63      & 15.75       & 46.32          & 34.82  \\
      %  & ATTA-1  &35.18          & 13.28     & 26.77      & 15.73       & 39.10          & 26.32  
       
         ALP~\cite{58kannan2018adversarial}    &46.58          & 15.34     & 36.51      & 18.43       & 45.94          & 33.07  \\ 
        \rowcolor{mygray}
        +\textbf{Ours}     & 46.11 & \textbf{17.25} & \textbf{38.33} & \textbf{21.43}  & 45.74    & \textbf{34.88}\\
         Trades~\cite{Zhang2019tradeoff} &42.49          & {12.97}     & 31.94      & 14.78       & 42.00          & 28.22  \\ 
        \rowcolor{mygray}
         +\textbf{Ours}    & \textbf{43.01} & \textbf{16.99} & \textbf{35.42} & \textbf{19.84}  & \textbf{42.54 }   & \textbf{31.80} \\
           CCG\cite{tack2021consistency}    &42.47          & 9.37      & 32.63      & 13.54       & 45.09          & 30.85  \\ 
        \rowcolor{mygray}
        +\textbf{Ours}    & \textbf{46.51} & \textbf{11.96} & \textbf{35.75} & \textbf{14.93}  & \textbf{46.50}    & \textbf{31.60} \\
        %& AWP     &46.24          & 14.27     & 38.53      & 19.72       & 47.61          & 36.10  \\ 
        \hlinew{1pt} 
        \end{tabular}
    \end{table*}
 
\newpage   
\noindent\textbf{t-SNE embeddings.} To validate the superiority of our proposed FAAT framework, we compare the feature distributions in models trained with different methods. We visualize the feature distribution using t-SNE embedding and show the results in Figure\,\ref{fig:distribution}. As we can see, both natural and adversarial data distribution learned by ALP has much larger intra-class distance and small inter-class distance, which will lead to lower accuracy in natural and adversarial data. ATTA and Trades enlarge the inter-class distance and make the decision boundary easier to decide. As we incorporate the class knowledge into trade-off balance, our proposed FAAT demonstrates the highest intra-class compactness and largest inter-class margin. Also, the features from natural data and adversarial examples have similar distributions, which is consistent with our motivation.

\begin{figure*}[!h]
\centering
\includegraphics[width=0.95\linewidth]{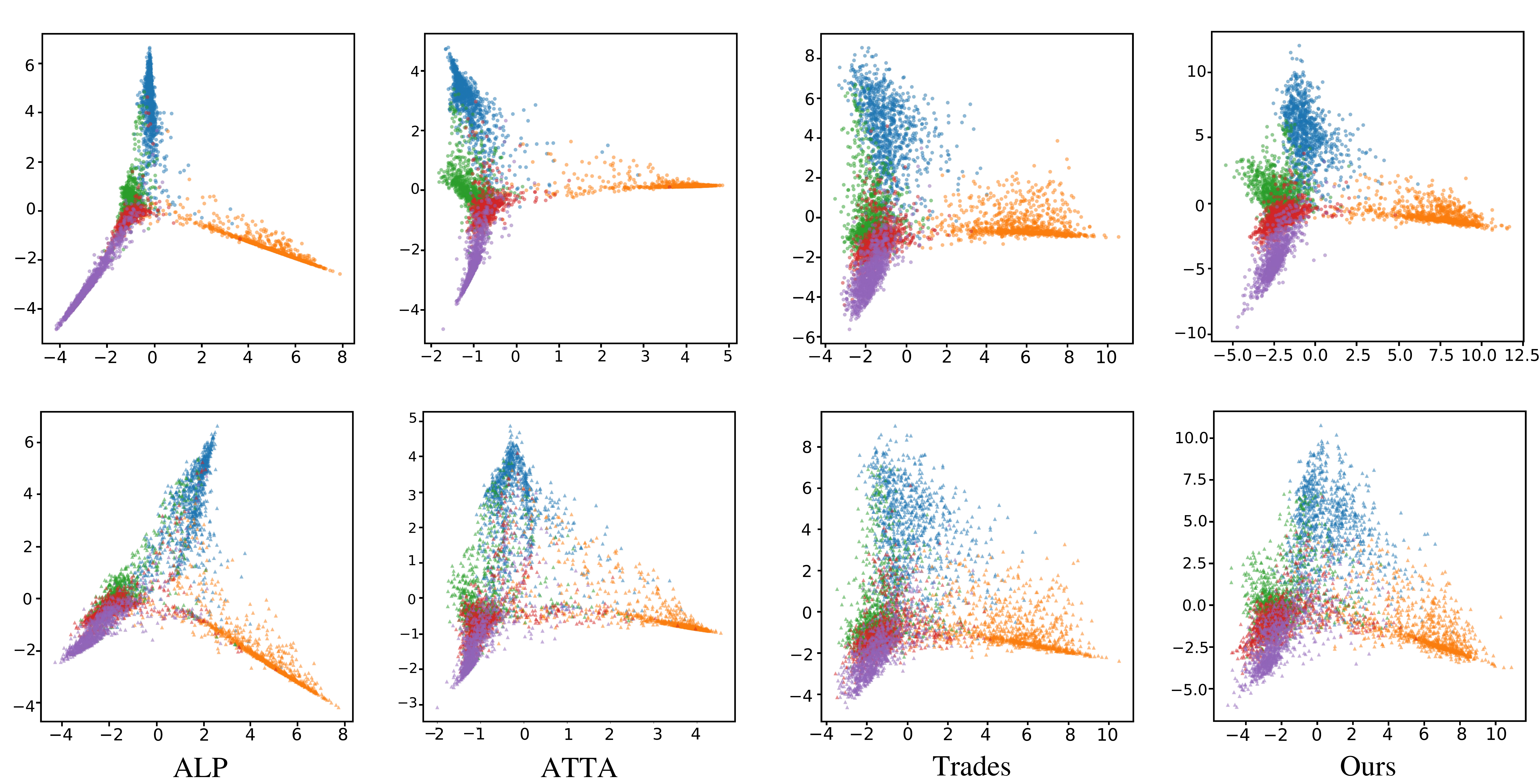}
    \caption{Visualization of feature distribution in WideResNet-34-10 model trained with different methods on CIFAR-10 dataset. Upper row: natural feature distribution. Below: adversarial feature distribution.}
\label{fig:distribution}
\end{figure*}

%-------------------------------------------------------------------------

\subsection{Experiments on Tiny-ImageNet}

Tiny-ImageNet\footnote{https://www.kaggle.com/c/tiny-imagenet} is a more complex dataset with more classes. We additionally conduct experiments on Tiny-ImageNet and report the results in Table\,\ref{tab:imagenet}. The experimental settings are the same with previous experiment on other datasets. As can be observed, the proposed FAAT framework achieves the best performance against both white-box attacks and black-box attacks. Moreover, incorporating FAAT into adversarial training can maintain the trade-off between the natural and robust accuracy as the FAAT also demonstrates the highest natural accuracy. These results not only imply that the FAAT framework can improve the overall performance of adversarial training against various attacks but also show that our method is more effective and feasible for large-scale datasets where vast classes of images are involved.
  \begin{table*}[!h]
        \centering
       % \fontsize{9}{11}\selectfont
        \caption{Natural and robust accuracy(\%) of WideResNet-34-10 trained on Tiny-ImageNet dataset using different adversarial training methods.}
        \label{tab:imagenet}
        \begin{tabular}{llcccccc}
        \hlinew{1pt}
        \multirow{2}{*}{{Method}} & \multirow{2}{*}{{Natural}}
        & \multicolumn{3}{c}{White-box} & \multicolumn{3}{c}{Black-box} \\ \cline{3-8}  
    &   & PGD-20          & PGD-50          & CW$_{\infty}$    & PGD-20          & PGD-50          & CW$_{\infty}$      \\ \hlinew{1pt}
    Standard~\cite{madry2018towards}     &   45.34   &  21.62        &  21.37         &  20.70     &  30.02     &  29.98        & 37.25 \\
         %ALP~\cite{58kannan2018adversarial}     &   45.20   &  22.37        &  22.21         &  21.50     &  32.22     &  32.08        & 45.10 \\
         Trades~\cite{Zhang2019tradeoff}     &   48.29   &  19.93        &  19.35         &  19.21     &  28.79     &  28.32        & 35.10 \\
        %& FREE (m=8) & 64.66          & 15.94           & 51.86       & 26.74       & 64.15          & 46.56 \\
       % & ATTA-1     & 67.99          & 16.95           & 58.85       & 28.50       & 72.47          & 57.97 \\
       
        %& AWP        & 73.74          & 25.49           & 61.27       & 29.62       & 72.20          & 53.86 \\ \rowcolor{mygray} 
     \textbf{Ours}  &  52.04  &  22.56      &    22.45        &  21.71      &   32.36   &     32.27    &   45.16 \\ \hlinew{1pt}
                    
        \end{tabular}
    \end{table*}

%\section{Comparison with AWP}

\end{document}